\documentclass[letterpaper]{article} 
\usepackage[]{aaai24}  
\usepackage{times}  
\usepackage{helvet}  
\usepackage{courier}  
\usepackage[hyphens]{url}  
\usepackage{graphicx} 
\usepackage{natbib}  
\usepackage{caption} 
\usepackage{algorithm}
\usepackage{algorithmic}
\usepackage{xcolor}
\usepackage{booktabs}
\usepackage{multirow}
\usepackage{newfloat}
\usepackage{listings}
\DeclareCaptionStyle{ruled}{labelfont=normalfont,labelsep=colon,strut=off} 
\floatstyle{ruled}
\newfloat{listing}{tb}{lst}{}
\floatname{listing}{Listing}
\usepackage{optidef}

\DeclareMathOperator*{\minimize}{minimize}

\title{Interactive Mars Image Content-Based Search with \\ Interpretable Machine Learning}
\author {
    Bhavan Vasu\textsuperscript{\rm 1, \rm 2},
    Steven Lu\textsuperscript{\rm 1},
    Emily Dunkel\textsuperscript{\rm 1},
    Kiri L. Wagstaff\textsuperscript{\rm 1, \rm 2}\\
    Kevin Grimes\textsuperscript{\rm 1},
    Michael McAuley\textsuperscript{\rm 1}
}
\affiliations {
    \textsuperscript{\rm 1} 
    Jet Propulsion Laboratory, California Institute of Technology, 4800 Oak Grove Drive, Pasadena, CA
    91109-8099, USA \\
    \textsuperscript{\rm 2} Oregon State University, Corvallis, OR 97331, USA\\
    vasub@oregonstate.edu, \{you.lu, emily.dunkel, kevin.m.grimes, michael.mcauley\}@jpl.nasa.gov, wkiri@wkiri.com
}

\begin{document}

\maketitle

\begin{abstract}
The NASA Planetary Data System (PDS) hosts millions of images of planets, moons, and other bodies collected throughout many missions. The ever-expanding nature of data and user engagement demands an interpretable content classification system to support scientific discovery and individual curiosity. In this paper, we leverage a prototype-based architecture to enable users to understand and validate the evidence used by a classifier trained on images from the  Mars Science Laboratory (MSL) Curiosity rover mission.
In addition to providing explanations, we investigate the diversity and correctness of evidence used by the content-based classifier. The work presented in this paper will be deployed on the PDS Image Atlas, replacing its non-interpretable counterpart.
\end{abstract}

\section{Introduction}
The PDS Cartography and Imaging Sciences Node is the curator of NASA's primary digital image collection spanning past, present, and future planetary missions. The PDS Imaging Node provides access to this data archive via the PDS Image Atlas. Due to the data archive's ever-growing nature
, manually searching through tens of millions of images to find data products of interest is infeasible. With an ever-growing list of missions, future releases of the Atlas will accommodate a reliable, interpretable content-based classification system that aims to provide a three-fold benefit. 
 
\par Firstly, an interpretable content-based classification system will validate the evidence used by the content-based classifier and ensure the right visual cues are being used by the classifier. Secondly, such an interpretable model will help bridge the gap between the mental model of Atlas users and planetary image classifiers. 
Finally, identifying erroneous evidence through user adjudications can establish a feedback loop from Atlas users to data scientists regarding the quality of evidence used by the planetary image classifiers.  Users searching Atlas will have the ability to interactively identify relevant images. Establishing such a feedback loop is vital for improving classifier performance while enabling users to play an active role, increasing user engagement and understanding.
\par In this paper, we employ explanations from case-based reasoning approaches that identify the evidence from the training set used to classify a test image. 

 We demonstrate an interpretable content-based image search system by leveraging the prototypical architecture proposed by~\citet{chen2019looks}. In addition to building upon the existing work, we extend it by evaluating the diversity and correctness of prototypes resulting from a classifier trained on imagery from Mars. Based on observations we made during the evaluation, our contribution was the incorporation of a diversity-enhancing term to the original work by \citet{chen2019looks}, which notably amplified the diversity of evidence utilized and subsequently enhanced performance. We describe our plan to deploy the system by replacing the non-interpretable counterpart currently hosted on the Atlas with the interpretable content-based classifier proposed in this paper. The MSL surface dataset \footnote{https://zenodo.org/record/4033453} used in this paper was 
published by \citet{wagstaff2021mars}.

\section{Related Work}
Interpretable content-based classification has appeared in the literature multiple times based on the classifier under investigation for content classification \cite{nauck1999obtaining, 718510, vasu2021explainable}.  
Further improvements in the wider field of interpreting machine learning decisions were achieved with the introduction of increasingly complicated and opaque classifiers.
Explanations generally appear under different taxonomies such as white-box vs. black-box, inherently interpretable vs. post-hoc, and neuron vs. primary vs. layer attribution methods \cite{lucas2022rsi}. White vs. black-box categorizes methods based on whether they leverage internal classifier structure to generate explanations. Inherently interpretable vs. post-hoc categorizes them based on the model naturally providing explanations vs. using an explanation method after model development. 

\par The introduction of deep models sacrificed interpretability in favor of improved performance and automatic feature extraction. Interpreting deep models has gained traction in recent years due to the large-scale deployment of deep models. One of the first works to interpret deep convolution neural network (CNN), proposed by  \citet{zeiler2014visualizing}, deals with understanding activations and the 
internal operations of CNNs by looking for patches that maximize a neuron activation. \citet{zhou2016learning} leveraged the presence of a global average pooling layer to backtrack and combine activations with the strongest connection to a particular class to produce local explanations in the form of saliency maps. 

Several works were also inspired by the gradient-based approach known as GradCAM proposed by \citet{selvaraju2017grad}. More recently \citet{khorram2021igos++} used integrated gradient and an optimization paradigm to obtain explanations. Interpretability can also come in the form of attention mechanisms. \citet{zheng2017learning, zhang2014part, DBLP:journals/corr/abs-1806-07421} offer some insight into regions of the image attended, indicating the "where" but fails to address "why" a region of the image was paid attention to. 
\par Despite the vast body of work in post-hoc explanations \cite{selvaraju2017grad, DBLP:journals/corr/abs-1806-07421, zheng2017learning}, we primarily focus on leveraging an inherently interpretable deep model such as ProtoPNet~\cite{chen2019looks} due to shortcomings of post-hoc approaches highlighted by~\citet{adebayo2018sanity}, ~\citet{rudin2019stop}, and ~\citet{lakkaraju2020fool}. Methods by \citet{selvaraju2017grad} are unreliable in the presence of repetitive patterns spread across a larger portion of the image, such as rocks or terrain on Mars. More importantly, we chose ProtoPNet because we expected its explanations to be more intuitive for users of the Atlas who may not be ML experts.
Case-based reasoning work by \citet{kolodner1992introduction}  uses previously created prototypes, limiting adaptivity, while the work by \citet{aamodt1994case} enhances adaptability through a four-phase approach involving both past and new prototypes. In contrast, the Prototypical Part Network (ProtoPNet) proposed by ~\citet{chen2019looks} creates new prototypes by optimizing network weights, focusing on minimizing the distance between prototypes and class instances in a high-dimensional space.
This paper leverages the architecture and training routine proposed by~\citet{chen2019looks}. We quantify the evidence learned by the deep models in terms of diversity and correctness of evidence as opposed to quantifying them through agreement~\cite{bau2017network, bau2018gan} with a region of the image as the latter requires a large volume of fine-grained annotations.

\section{Prototypical Part Network}
The ProtoPNet architecture proposed by~\citet{chen2019looks} uses a standard deep network for feature extraction and introduces a prototypical layer that learns a pre-defined number of prototypes or exemplars that best represent each image class. The fully connected layer after the prototypical layer represents the contribution of each learned prototype to the final decision. Therefore, the network avoids problems such as unreliability due to inaccuracies in understanding a model's decision-making process, described by ~\citet{rudin2019stop} by only allowing information to flow to the final classification layer through the prototypical layer. 
Once regions of images most similar to the learned prototypes are found through similarity matching, they can be visualized as a bounding box by using a threshold at 95\% of maximum similarity. The similarity score represents the strength of the prototype match, while the weights of the fully connected layer represent its contribution to a class during training. Above all the highlighted benefits, ProtoPNet learns part-based associations with no additional region-based annotation, which allows the developer to define a finer subclass.

\par
\begin{figure}
    \centering
\includegraphics[width=0.45\textwidth]{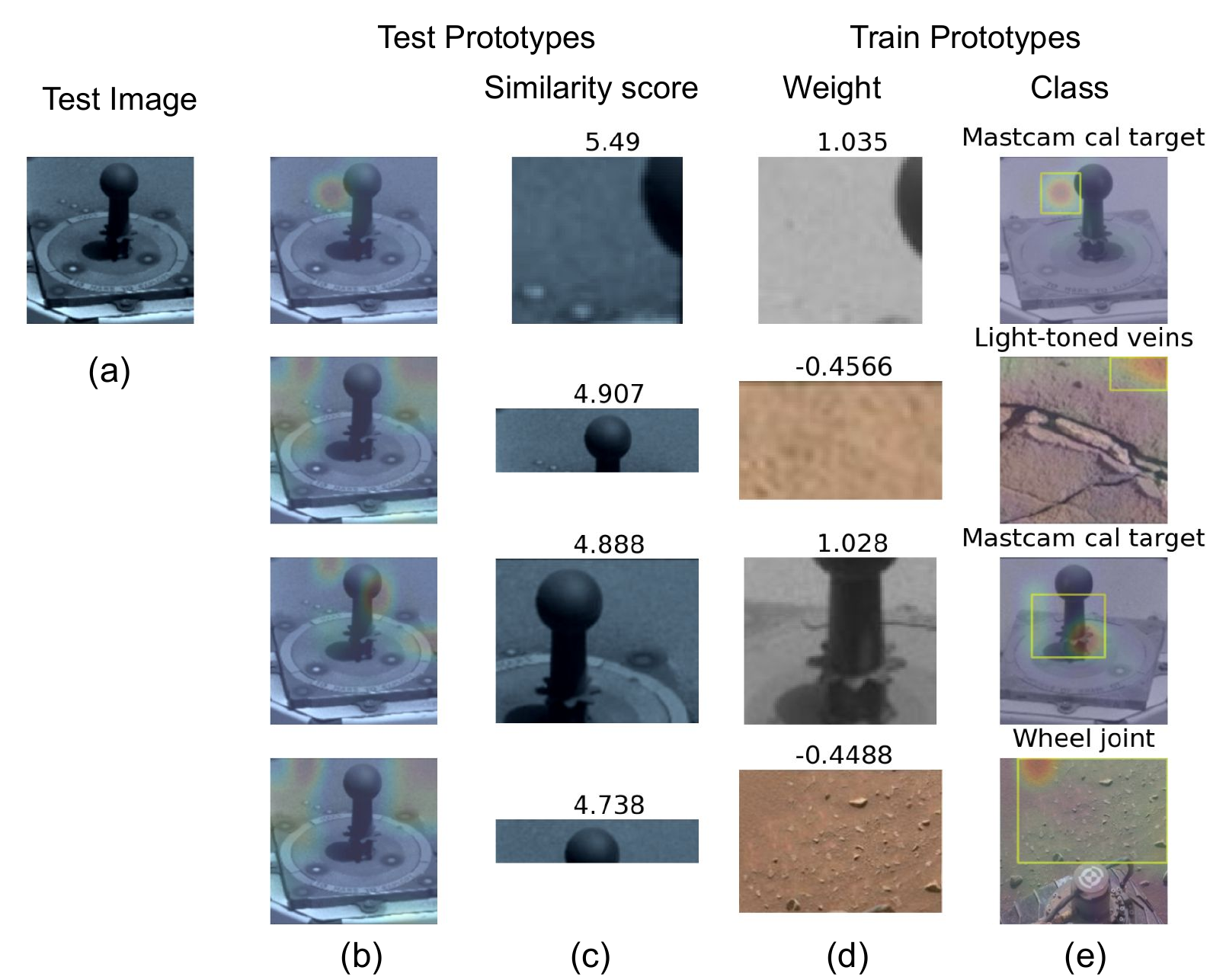}
    \caption{Qualitative example of the top-4  most visually similar prototypes for class \textit{Mastcam cal target} from the MSL surface dataset. Column (a) is the test image, (b) shows the same image overlayed with a heatmap showing regions most activated by the prototype learned during training followed by (c) showing a cropped version of the heatmap after threshold with the similarity score. (d) shows the cropped regions of heatmaps from Column (e). Column (e) shows the training images overlayed with regions obtained after prototypes projection on the training set and .  The evidence looks coherent across both training and testing prototypes i.e., column (c) and (d) when the weights are positive.}
    \label{fig:mastcam_res}
\end{figure}

 In the following sections, we present results for both a VGG19 and ResNet18 backbone on the MSL surface  dataset. The MSL surface dataset contains visual features observed on the surface of Mars by the Curiosity rover. Figure \ref{fig:mastcam_res} (discussed in more detail later) shows an example ProtoPNet output for the MSL surface dataset image belonging to class \textit{Mastcam cal target} used for calibrating the Mastcam instrument on board the rover. We use ten prototypes per class for all experiments in the rest of the paper. We use the weights pre-trained from ImageNet \cite{ILSVRC15} data set to initialize all our feature extractor backbones.

\section{Proto-MSLNet}

\par In this section, we elucidate the training process of the content-based classifier utilizing the MSL surface dataset, herein referred to as MSLNet. The currently deployed version of MSLNet on the Atlas classifies imagery received from the MSL Curiosity rover. We plan to replace MSLNet with an interpretable version we call Proto-MSLNet, which is constructed by training a prototypical and fully connected layer on top of an off-the-shelf feature extractor backbone. Due to the benefits of interpretable models outlined in the earlier sections, we plan to deploy Proto-MSLNet in spite of a slight drop in test accuracy. Based on our observations while using current approaches, we noticed a lack of diversity among prototypes, i.e., prototypes often look visually similar or come from the same training image. Therefore, we modified the ProtoPNet training procedure to incorporate a diversity factor:
\begin{align*}
    \text{Div }  =  - \frac{1}{n}\sum_{x=1}^{n} &\minimize_z \max(z - p_j - \text{margin}, 0) \\
    \text{subject to  } &z \in \text{patches}(f(x_i))\\
    &j:p_j \in P_{y_i}
\end{align*}
where $n$ represents the total number of images under consideration. The image $x_i$ denotes the $i^{th}$ image in the dataset, and its corresponding feature representation is captured by $f(x_i)$ while $\text{patches}(f(x_i))$ constitutes all potential patches derived from $f(x_i)$. $p_j$ represents learned feature prototype corresponding to the class $y_i$ and $P_{y_i}$ is the complete set of prototypes for class $y_i$. Lastly, the margin term is a predefined threshold to eliminate trivial differences among prototypes. 
The total loss $L$ used to train Proto-MSLNet is
\begin{align*}
   L =  \minimize_{\mathbf{P}, w_{conv}} \frac{1}{n} \sum_{i=1}^n \text{CrsEnt}(h \circ g_{\mathbf{P}} \circ f(x_i), y_i) \\+ \lambda_1 \text{Clst}+ \lambda_2 \text{Sep} + \lambda_3 \text{Div}
\end{align*}
where the cross entropy loss (CrsEnt) penalizes misclassification on the training data. The clustering cost (Clst) promotes the presence of a latent patch in each training image that is proximate to at least one prototype from its class. Conversely, by minimizing the separation cost (Sep), it is encouraged that every latent patch of a training image remains distant from prototypes that do not belong to its class. While Clst brings together prototypes of the same class and Sep promotes inter-class separation in prototypes, there is no condition promoting intra-class diversity in prototypes. To address this issue, we introduce the diversity cost (Div).  Note, $\lambda_1, \lambda_2, \text{and } \lambda_3$ are hyperparameters used to control the influence of each cost term, $g_{\mathbf{P}}$ is the prototype layer, and $h$ is a fully connected layer. To visualize the prototypes as images we project the prototypes onto the training set as shown below:
\begin{align*}
    p_j \leftarrow arg \min_{z \in \mathcal{Z}_j} \lVert z - p_j \lVert_2
\end{align*}
Where $\mathcal{Z}_j$ represents the set of all training patches and $p_j$ is the prototypes found while computing $L$. Similarly, we can also visualize the test prototypes by replacing $\mathcal{Z}_j$ with the test set.

\subsection{Data Set}
\begin{figure}[t]
    \centering    
    \includegraphics[width=0.4\textwidth]{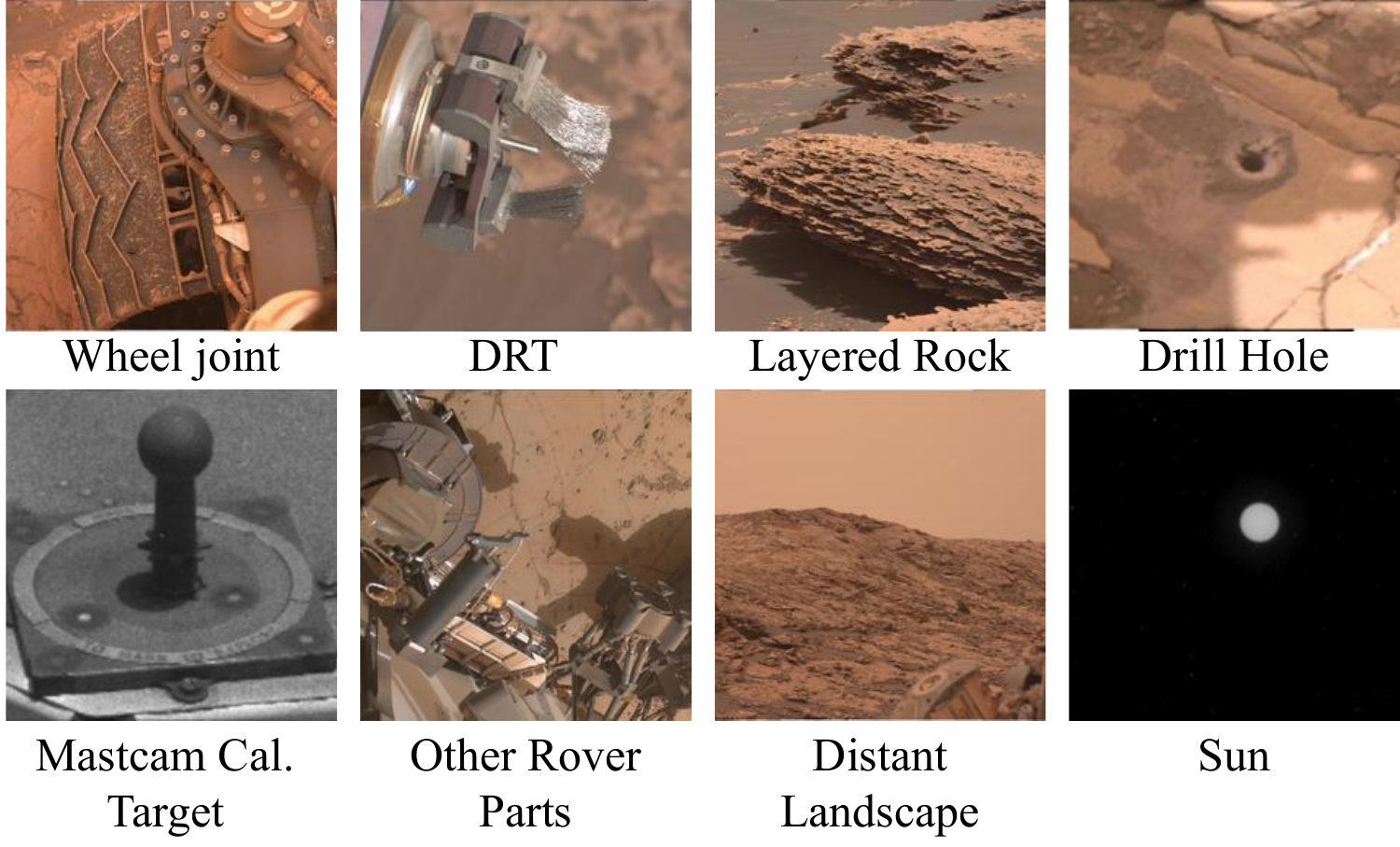}
    \caption{Figure showing representative examples from eight classes of the MSL surface Data Set. DRT refers to the Dust Removal Tool aboard the Curiosity rover.}
    \label{fig:data_msl}
\end{figure}

The MSL surface dataset  was collected by the Mast Camera (Mastcam) and Mars Hand Lens Imager (MAHLI) instruments on the MSL Curiosity rover spanning 19 classes of interest such as Dust Removal Tool (DRT), Sun, Night Sky, Wheels, Wheel joints, and Wheel tracks. We augment the dataset according to the realistic variations for each instrument: the images from the rotatable platform MAHLI are rotated by 90, 180, and 270 degrees with horizontal and vertical flipping, and the images from the fixed platform Mastcam are only horizontally flipped. Figure \ref{fig:data_msl} shows representative images from eight different classes along with the result of data augmentation on an image from class \textit{Wheel} acquired using MAHLI.

\begin{figure*}[t]
    \centering
    \includegraphics[width=0.70\textwidth]{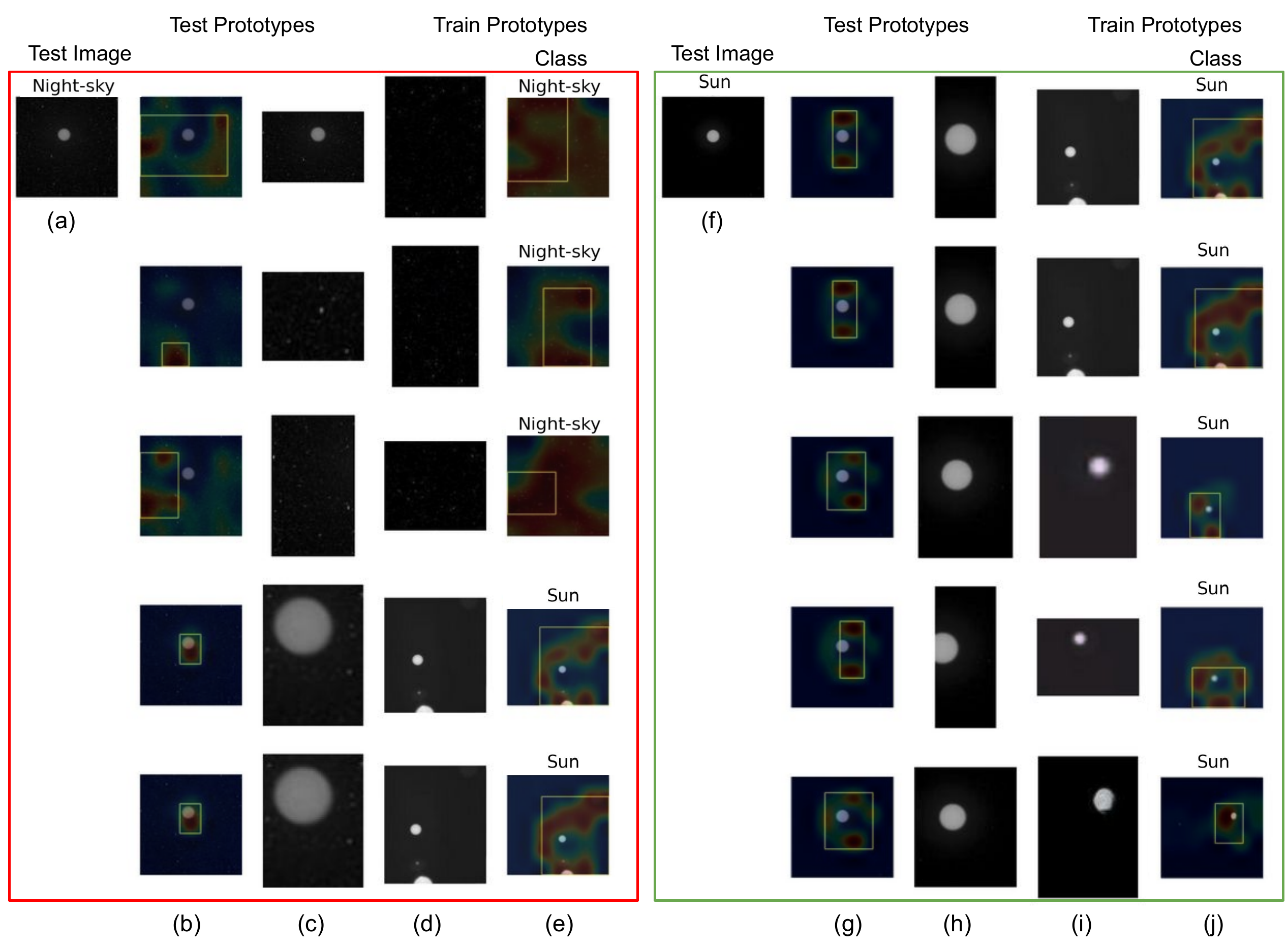}
    \caption{Explanation for two images from class \textit{Sun} showing the difference between evidence when the image is misclassified as \textit{Night Sky} (red, left) vs. when it is classified correctly as \textit{Sun} (green, right)  from a VGG19 backbone. The meaning of the columns is identical to Figure \ref{fig:mastcam_res} where (a - e) represents output for the test image in (a) and (f - j) represents output for the test image in (f). Note the prototypes are ordered from most similar to least.}
    \label{fig:sun_res}
\end{figure*}

\fontsize{10pt}{11pt}\selectfont

\begin{table*}[t]
\centering
\caption{Train, validation, and test accuracy along with threshold accuracy and abstention rate of different deep networks trained on the MSL surface dataset. MSLNet is a regular deep CNN; P-MSLNet is its prototypical version. Note that the performance metrics reported in this table do not incorporate the diversity loss; they solely aim to elucidate the influence of employing a prototypical layer.}
\label{tab:msl_acc}
\footnotesize
\setlength{\tabcolsep}{3.5pt} 
\begin{tabular}{@{} l ccc ccc ccc @{}}
\toprule
\multirow{3}{*}{\textbf{Model Configuration}} & \multicolumn{3}{c}{\textbf{Train (n=5{,}920)}} & \multicolumn{3}{c}{\textbf{Validation (n=300)}} & \multicolumn{3}{c}{\textbf{Test (n=600)}} \\
\cmidrule(lr){2-4} \cmidrule(lr){5-7} \cmidrule(lr){8-10}
 & \textbf{Acc} & \textbf{Acc$_{0.9}$} & \textbf{Abst} & \textbf{Acc} & \textbf{Acc$_{0.9}$} & \textbf{Abst} & \textbf{Acc} & \textbf{Acc$_{0.9}$} & \textbf{Abst} \\
 & \textbf{(\%)} & \textbf{(\%)} & \textbf{(\%)} & \textbf{(\%)} & \textbf{(\%)} & \textbf{(\%)} & \textbf{(\%)} & \textbf{(\%)} & \textbf{(\%)} \\
\midrule
Most common (baseline) & 26.3 & - & - & 24.7 & - & - & 31.2 & - & - \\
\midrule
\textbf{VGG19} \\
\quad MSLNet & 99.4 & 99.8 & 1.5 & 81.6 & 85.6 & 14.3 & 81.3 & 85.6 & 12.0 \\
\quad + P-MSLNet & 99.3 & 99.8 & 4.8 & 77.3 & 83.1 & 21.0 & 75.1 & 82.5 & 19.8 \\
\quad + P-MSLNet - TempCal & 99.3 & 99.7 & 2.5 & 77.3 & 80.1 & 14.6 & 75.1 & 94.5 & 63.3 \\
\quad + P-MSLNet - VectorCal & 99.3 & 99.8 & 2.4 & 81.3 & 94.8 & 55.0 & 79.6 & 94.9 & 53.8 \\
\midrule
\textbf{ResNet18} \\
\quad MSLNet & 100.0 & 100.0 & 0.0 & 83.0 & 87.6 & 21.6 & 79.5 & 86.3 & 15.5 \\
\quad + P-MSLNet & 96.4 & 98.6 & 7.3 & 76.0 & 84.5 & 24.3 & 74.8 & 83.0 & 24.5 \\
\quad + P-MSLNet - TempCal & 96.4 & 98.9 & 10.6 & 76.0 & 94.9 & 60.6 & 74.8 & 91.6 & 66.1 \\
\quad + P-MSLNet - VectorCal & 96.5 & 98.8 & 10.4 & 77.6 & 95.2 & 57.7 & 76.8 & 92.2 & 61.1 \\
\bottomrule
\end{tabular}
\end{table*}

\begin{figure}[t!]
    \centering
    \includegraphics[width=0.4\textwidth]{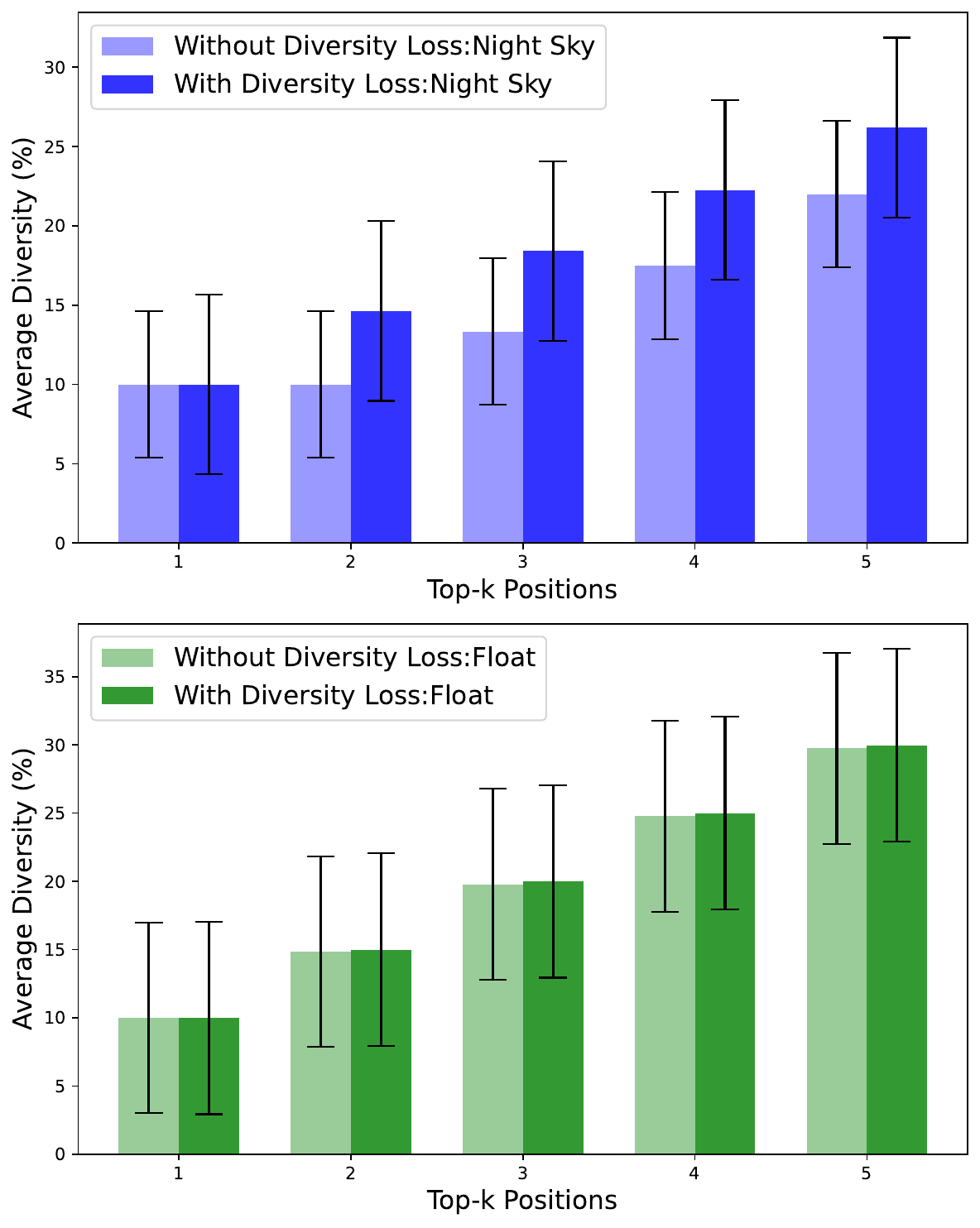}
    \caption{Comparison of the average prototype diversity over 100 prototypes for the most and least diverse classes, plotted against the position of the prototype based on the order of evidence (sorted based on importance), denoted as $k$, used for classifying MSL surface data. Class  \textit{Night Sky} sees significant improvement while class \textit{Float Rock} has no improvement in diversity from the inclusion of the diversity loss term.}
    \label{fig:div_msl}
\end{figure}
\begin{figure}[t!]
    \centering    \includegraphics[width=0.43\textwidth]{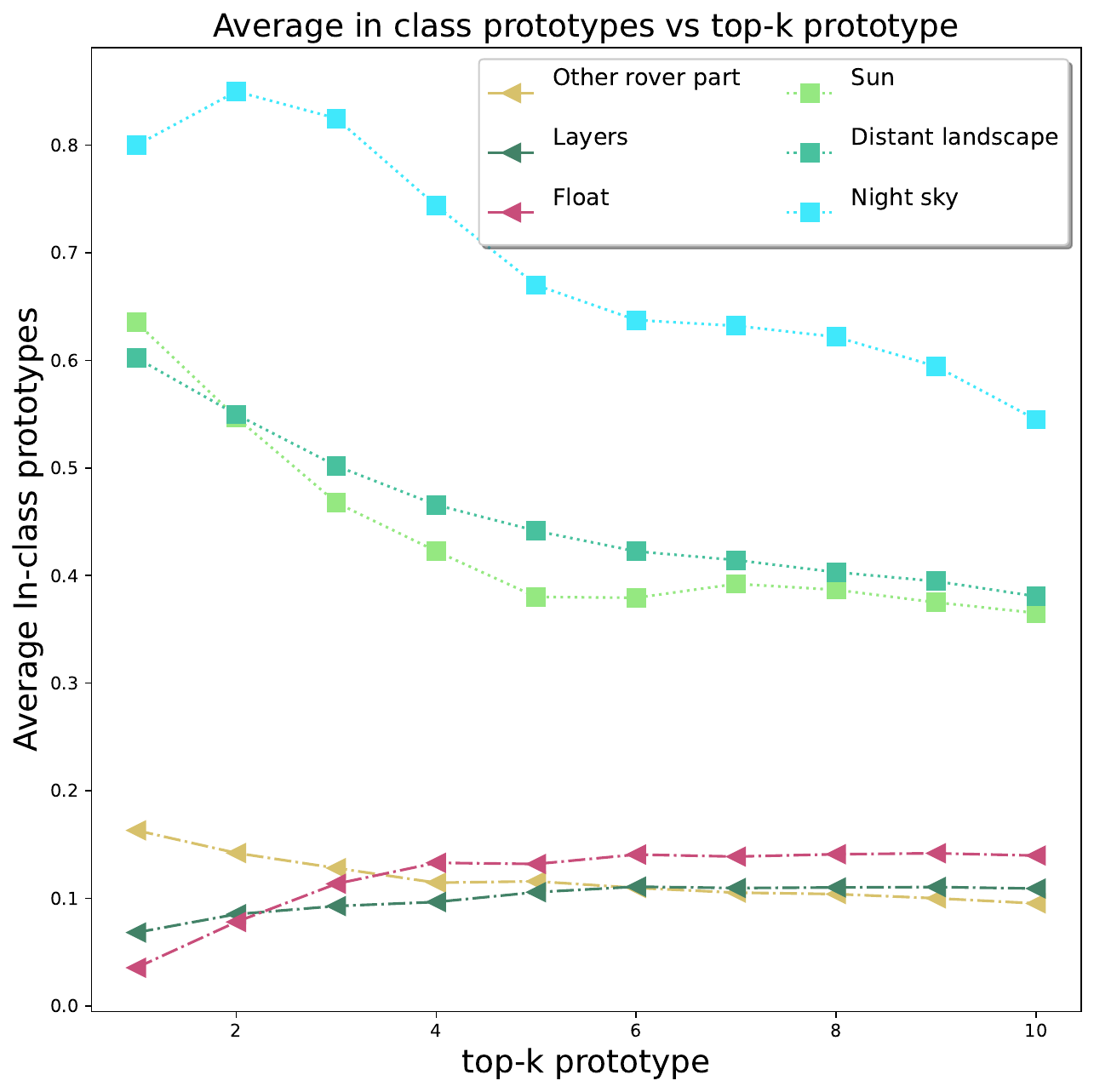}
    \caption{Average number of In-class prototypes for top-3 (square) and bottom-3 (triangle) most correct classes vs position of prototype in the order of evidence used $k$ for classifying MSL surface test data by the VGG19 version of ProtoPNet.}
    \label{fig:inclass_msl}
\end{figure}

\subsection{Experimental Setup}
During training, we use a learning rate of 1e-4 for the first 100 epochs followed by a learning rate of 1e-5 for 100 epochs and select the model with the best validation accuracy. Note the training process is split into a gradient update at each epoch and a projection stage every 5 epochs. The hyperparameters $\lambda_1, \lambda_2, \text{and } \lambda_3$ are assigned values of $0.8, 0.08, \text{and } 0.04$ respectively. These values were determined through an empirical evaluation across five distinct trials. A more rigorous parameter search will be undertaken in preparation for deployment. We employ a sol-based split as proposed by \citet{wagstaff2021mars} to evaluate generalization in a realistic setting, wherein training occurs on past data and validation/testing on future data. The term "sol" here refers to a measure of one Mars day. Note that the sol-based split reveals the temporal label shift between train and validation/test set, reflected in the gap between their accuracy across both deep CNNs reported in Table \ref{tab:msl_acc}. \citet{wagstaff2021mars} provide a full description of the dataset generation process with detailed class distributions. A batch size of 80 was used in combination with an Adam optimizer~\cite{kingma2014adam} to train all the classifiers. All classifiers were trained on a single V100 GPU facilitated by the Texas Advanced Computing Center (TACC).

\section{Results and Analysis}

We will discuss our quantitative comparison, followed by a discussion on qualitative results and an analysis of the correctness and diversity of prototypes. 
\subsection{Quantitative Analysis}
\par Table \ref{tab:msl_acc} reports train, validation, and test set accuracy on the MSL surface data set with and without a prototypical layer. Note, that the number of images in each dataset split is indicated in the heading. In Table \ref{tab:msl_acc}, we also report the train, validation, and test set accuracy of all classifiers for only images that were classified with a confidence higher than 90\% under column ``Acc(0.9)" and the percentage of examples under the 90\% confidence threshold (omitted from the Atlas display) listed as ``Abst Rate". This is crucial as only classification results with a confidence level of 0.9 or greater are delivered to the Atlas. We also report the baseline accuracy when predicting the majority class under ``Most Common", due to the imbalanced nature of our dataset. At an abstention rate less than $20\%$, the ResNet18 architecture fine-tuned on the MSL surface dataset yields the highest ``Acc(0.9)" of 86.39\% with an abstention rate of 15.5\%. We observe a drop of $\sim 2.5\%$ in ``Acc(0.9)" and a 9 percentage points rise (from $15.5\%$ to $24.5\%$) in abstention rate compared to its uninterpretable counterpart. A slight drop in accuracy is expected as the addition of the prototypical layer constraints the information passed on to the final layers to be the prototypes. The work by \citet{chen2019looks} suggests an ensemble of different backbone networks to further close the performance gap between a network with and without a prototypical layer. Furthermore, in future research endeavors, we intend to systematically examine the impact of employing a prototypical layer on the abstention rate, particularly following the application of model calibration techniques\cite{guo2017calibration}. 

\subsection{Qualitative Analysis}
\par Figure \ref{fig:mastcam_res} displays results for a sample image from the \textit{Mastcam cal target} class. The calibration target, as seen in column (d), rows 1 and 3, emerges as the predominant prototype for classification. In contrast, the prototypes representing the background, located in column (d), rows 2 and 4, are secondary in influence. Notably, a significant portion of training images from the \textit{Mastcam cal target} class, captured by the rover's integrated cameras, showcased a rocky backdrop surrounding the calibration target. This observation indicates that both the calibration target and its background contribute as evidence. Specifically, the background operates as negative evidence, evidenced by its negative weights of -0.45 and -0.44 in column (d), rows 2 and 4, respectively, guiding the accurate classification of the \textit{Mastcam cal target}.

\par To demonstrate the agreement between the correctness of prototypes and classification results, we present prototypes from two different test images from class \textit{Sun} in Figure \ref{fig:sun_res}. The evidence set marked in red(left) explains what led to misclassifying the image as class \text{Night Sky} while green(right) indicates a true positive. The top-3 prototypes in the incorrect classification focus on the emptiness of the night sky. This behavior might be a result of some training images from class \textit{Night Sky} having similar features as class \textit{Sun}, with both classes sharing visual attributes like the empty sky. On the other hand, when the image was classified correctly, the top evidence highly correlates to what a human might consider salient of the Sun, i.e., a bright, round shape. Therefore, using an interpretable content classification system can provide a deeper insight into the classifier's decision when compared to its un-interpretable counterpart.

\subsection{Diversity and In-class Prototypes}
\par To understand the nature of evidence used globally across classes, we present a quantitative evaluation of prototypes for their correctness and diversity. Evaluating the correctness of a prototype will help us to identify classes that use evidence from other classes, i.e., classes sharing visual attributes. In evaluating the diversity of prototypes learned during training, we aim to understand the generalizability of each class from the deep network's perspective. This is based on the assumption that a more diverse set of prototypes for a class enhances its ability to generalize. With our experimental setup, we noticed a 2 percentage point rise in test accuracy and no significant drop in train accuracy by introducing the diversity loss. Figure \ref{fig:div_msl} shows a plot of the average number of unique training images from which the most similar prototypes for correctly classified test images come vs. the position of prototype $k \in [1, 5 ]$. Ideally, we expect a linear growth rate when every prototype comes from a different image. Figure \ref{fig:div_msl} reports the results obtained using the VGG19 version of the ProtoPNet for the top-1 and bottom-1 most diverse classes. Class \textit{Night Sky} has the least diverse set of prototypes learned during training, with at most $\sim 4$ unique prototypes among ten prototypes. Class  \textit{Night Sky} sees significant improvement in diversity from the inclusion of the diversity loss term proposed in this paper. Note there is no change in diversity for class \textit{Float Rock} with the most diverse set of prototypes.

\par Finally, in Figure \ref{fig:inclass_msl}, we report the average number of in-class prototypes present in top-$k$ most similar prototypes, i.e., the evidence used for classification that comes from an image that is the same class as the test image. Examining in-class prototypes provides insights into how much an image class depends on other classes.
From Figure \ref{fig:inclass_msl}, it is evident that class \textit{Night Sky} seems to have the most in-class prototypes on average across all classes, with almost 80\% of the images having top-1 evidence coming from class \textit{Night Sky}. Similarly, classes \textit{Float Rock} and \textit{Layered Rock} seems to have the least number of in-class prototypes with only ~10\% - 15\% images containing in-class evidence indicating the lack of enough evidence during training between the two classes or the two classes have a lower inter-class variance. 

\section{Characterizing Prototypes}
\begin{figure}
    \centering
    \includegraphics[width=0.39\textwidth]{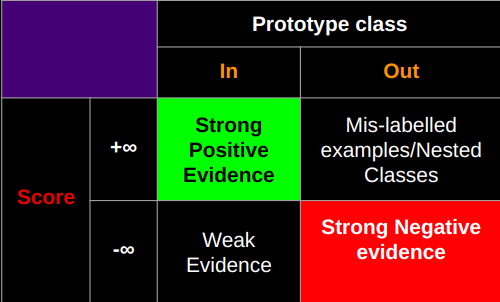}
    \caption{Characterizing prototypes according to their score and class distribution.}
    \label{fig:cfm}
\end{figure}

The benefit of explanations can be further explored by characterizing them, enabling data scientists to get deep insights into the prototype score and class distribution. Our approach involves a systematic classification of prototypes into distinct categories: Strong Positive Evidence, Strong Negative Evidence, Mislabeled or Nested examples, and Weak Evidence, as depicted in Figure \ref{fig:cfm}. This taxonomy not only aids in identifying mislabeled instances and nested classes with closely related contexts but also unveils the subtleties in class distinctions. Specifically, we report a notable overlap between the ``Wheel Joint" and ``Wheel" classes, primarily attributed to the camera's positioning and the restricted angles of rotation. Furthermore, in a separate dataset not covered in this manuscript, we identified a test image erroneously labeled as ``Impact Ejecta" which, in reality, encapsulated both ``Impact Ejecta" and ``Bright Dune," subsequently being associated with two ``Bright Dune" Prototypes.

\section{Deployment Plan}
The interpretable content-based classification system studied in this paper will be evaluated for deployment in place of its un-interpretable counterpart currently in operation on the PDS Image Atlas\footnote{https://pds-imaging.jpl.nasa.gov/search/}. Our deployment plan is as follows:
\begin{enumerate}
    \item We plan to create a comprehensive user study with different levels of information being displayed to the user to assess their relative merits. Some questions we hope to answer are: How many prototypes should be displayed to the user? How much feedback is the user willing to provide?
    \item Deploy the Proto-MSLNet classifier on MSL archives at the PDS Imaging Node after calibrating the posterior probabilities to inform decisions about which (high-confidence) images will be shown.
    \item  Driven by prior research in explanation visualization \cite{gunning2021darpa, vasu2021explainable}, we plan to modularize the cost of computing explanations to maintain user engagement while only providing explanations on-demand as explanations for simple decisions do not help the user. We plan to work with our software development and User Interface (UI)/User Experience (UX) design teams at the PDS Imaging Node to integrate the UX design shown in Figure \ref{fig:ui} on the Atlas.
    \begin{figure}[t]
    \centering
\includegraphics[width=0.47\textwidth]{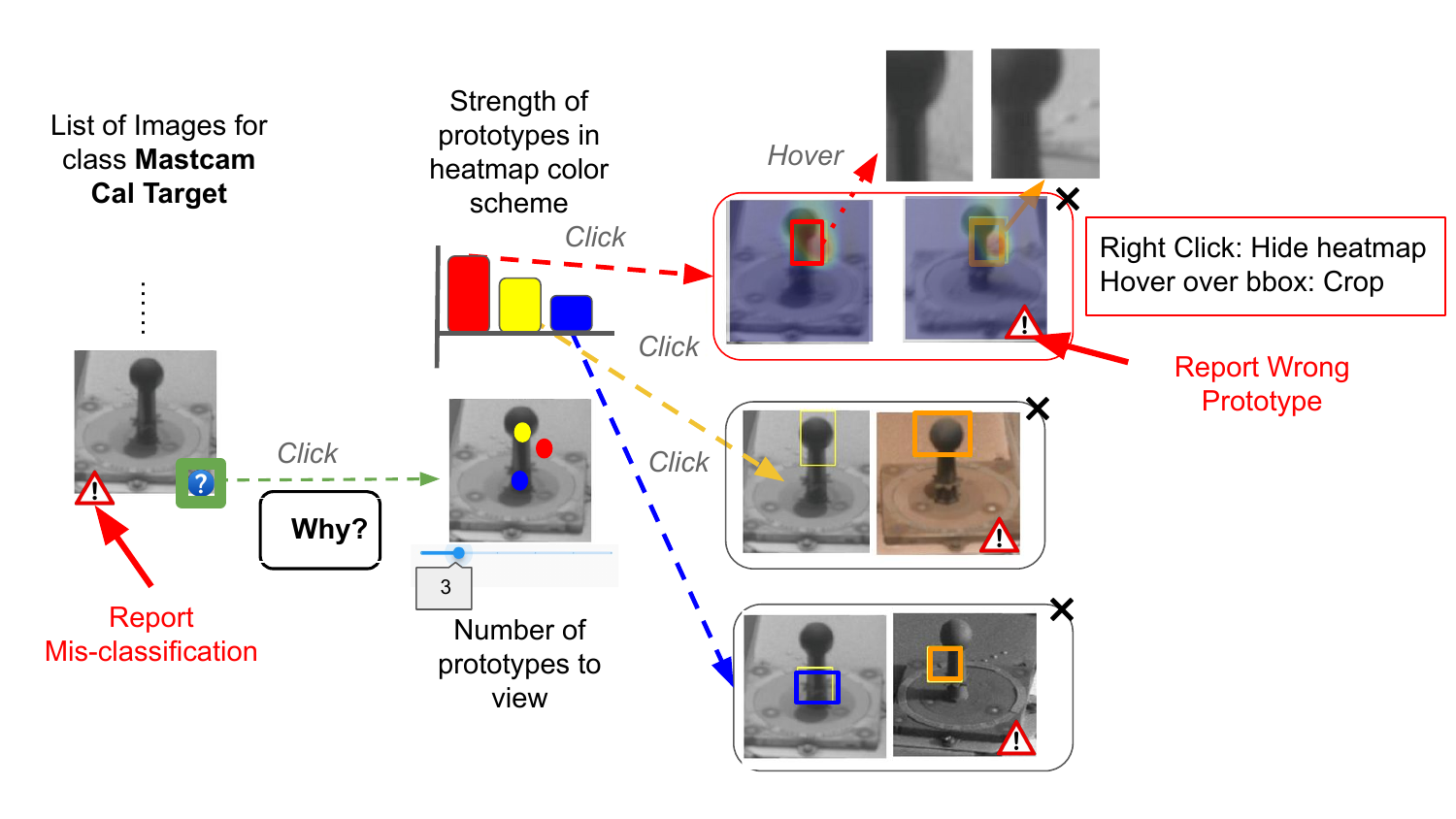}
    \caption{An illustration of user experience being considered for explanation visualization.}
    \label{fig:ui}
\end{figure}
    \item In addition to providing explanations, we would like to close the loop by allowing user to report misclassified images or erroneous evidence as shown in Figure \ref{fig:ui}. We would like to quantify and investigate methods to incorporate user feedback into model development. Our model refinement protocol incorporates a tripartite feedback integration strategy: (1) Database consolidation of misclassification feedback; (2) Analytical review to ascertain error causes; (3) Model enhancement based on prevalent error causes, including label rectification or training set expansion, prior to model retraining.
\end{enumerate}

\section{Conclusion}
In this paper, we present our plans to deploy an interpretable content-based search on the Planetary Data System (PDS) Image Atlas. Building upon prior research on prototypical networks, we introduced a novel enhancement through the integration of a diversity cost. We report the results for classifiers trained on Mars images acquired from instruments on the Curiosity rover. Firstly, we demonstrated that an inherently interpretable network could be trained for imagery from Mars with a minimal performance drop of $2.5$ percentage points with a $9$ percentage points rise in abstention rate compared to its non-interpretable counterpart. In addition to highlighting the benefits of having an interpretable system through qualitative examples, we also report quantitative metrics that help us judge the quality of evidence learned for an image class. The ability to provide evidence used for content classification lets us debug spurious evidence used by the classifier and paves the way for a feedback mechanism from its users to improve model performance. In future work, we plan to investigate the effect of such reporting mechanisms on overall system improvement. We also plan to investigate different visualizations of prototypes to understand user preferences. While this paper focused on results from classifiers trained on the MSL surface dataset, we intend to broaden our research to include other classifiers such as the Mars Reconnaissance Orbiter and Mars Exploration Rover present on the Atlas. Overall, the work presented in this paper aims to render content classification across all missions on the NASA PDS system transparent and interpretable accelerating scientific discovery and aiding individual curiosity.

\section{Acknowledgments}
We thank the PDS Imaging Node
for the continuing support of this work. We also thank
the numerous volunteers involved in labeling Mars images enabling the work in this paper. The computing resources were provided by the Texas Advanced Computing Center (TACC). This research was partially funded through the NSF grant CNS-1941892, the Industry-University Cooperative Research Center on Pervasive Personalized Intelligence, and was carried out at the Jet Propulsion Laboratory, California Institute of Technology, under a contract with the National Aeronautics and Space Administration. Copyright 2023. All rights reserved. 
\bibliography{aaai24}

\end{document}